\documentclass{article}
\PassOptionsToPackage{numbers,sort&compress}{natbib}
\usepackage[dblblindworkshop,final]{neurips_2026}
\usepackage[T1]{fontenc}
\usepackage[utf8]{inputenc}
\usepackage{amsmath,amssymb}
\usepackage{booktabs}
\usepackage{longtable}
\usepackage{wrapfig}
\usepackage{graphicx}
\usepackage{microtype}
\usepackage{xcolor}
\usepackage{url}
\usepackage[hidelinks]{hyperref}
\usepackage{tikz}
\usetikzlibrary{arrows.meta,positioning,calc}
\definecolor{kgteal}{HTML}{0E7C8B}
\definecolor{kgmuted}{HTML}{516470}
\definecolor{kgline}{HTML}{D6E0E6}
\definecolor{kgsoft}{HTML}{E4F1F3}
\hypersetup{pdftitle={When Copying Is Hard to Beat: A Negative Result in Clinical Assessment Forecasting},pdfauthor={},pdfsubject={},pdfkeywords={},pdfcreator={},pdfproducer={}}
\newif\ifauthornotes
\authornotesfalse
\definecolor{authorred}{HTML}{A63732}
\newcommand{\Acc}{\mathrm{Acc}}
\newcommand{\Accc}{\mathrm{Acc}_{\Delta}}
\newcommand{\FCR}{\mathrm{FCR}}
\newcommand{\ind}{\mathbb{1}}
\newcommand{\newstate}{\textit{new}}
\newcommand{\acutestate}{\textit{acute}}
\newcommand{\nonestate}{\textit{none}}
\title{When Copying Is Hard to Beat:\\A Negative Result in Clinical Assessment Forecasting}
\workshoptitle{Foundation Models for Temporal Systems: From Forecasting to World Modeling}
\author{%
  Marius Fracarolli\thanks{Also with HIDSS4Health --- Helmholtz Information
    \& Data Science School for Health, Karlsruhe/Heidelberg, Germany.} \\
  Department of Computational Linguistics\\
  Heidelberg University, Germany\\
  \texttt{fracarolli@cl.uni-heidelberg.de} \\
  \AND
  Stefan Riezler \\
  Department of Computational Linguistics \&\\
  Interdisciplinary Center for Scientific Computing (IWR)\\
  Heidelberg University, Germany\\
  \texttt{riezler@cl.uni-heidelberg.de} \\
}
\begin{document}
\maketitle

\begin{abstract}
We report a negative result in next-day clinical assessment forecasting: the evaluated models show
no consistent advantage over strong persistence-based baselines. Clinicians can view the previous
day's questionnaire answers, a plausible source of documentation persistence. Predicting 45 items
on 29{,}715 ICU day-pairs with patient-disjoint splits, a physiology-informed gated model improves
item-macro-F1 over a one-line clinical rule by $+0.0019$ $[+0.0006,+0.0031]$, while losing accuracy.
No macro-F1 advantage is established when the physiology window ends at the previous form's
reference time ($+0.0001$ $[-0.0006,+0.0008]$), or when scoring excludes the 68.5\% of cells derived
from parent answers; combining both changes yields a small disadvantage. Four checkpoints across three families
underperform persistence, and a fine-tuned one differs from it on only two cells.
A categorical read-out raises changed-cell accuracy eightfold, but a state-only read-out scores
higher overall than forecast-only decoding and matches a transition matrix. This negative result
highlights how the documentation process, input cutoff, observation mask and decoding rule shape
conclusions about forecasting recorded assessments.
\end{abstract}

\section{Introduction}
\label{sec:intro}
We report a negative result: the evaluated predictors do not consistently improve on strong simple
baselines for next-day clinical assessment forecasting. Clinicians completing the questionnaire can
view the previous day's answers, making persistence directly relevant to the documentation workflow.
Access to prior answers could contribute to label stability alongside genuine clinical stability;
the data cannot separate these effects. We therefore evaluate forecasts of recorded assessments,
whose temporal consistency may partly reflect how they are collected.

We predict the next day's values of 45 recorded items -- diagnostic judgments, infection
localization, organ dysfunction -- against persistence, a one-line organ-acuity rule, a per-item
transition matrix, supervised encoders, baseline-gated predictors and pretrained forecasters.
Persistence is a strong baseline~\citep{HewamalageETAL:23}, and labels of this kind have been used for sepsis
prediction~\citep{Schamoni2019}.

The strongest gated model gains $+0.0019$ item-macro-F1 over the rule, interval excluding zero, but
loses accuracy. Two task definitions qualify this gain. The \textbf{input cutoff}: the form carries a fixed daily
reference time, so the usual physiology window overlaps the interval the target form describes. The
\textbf{observation mask}: the instrument is conditional, and 68.5\% of scored cells are materialized
from a parent answer rather than entered. Under either, no macro-F1 advantage is established; under
both the model is significantly worse. Only that summary flips -- the gate loses accuracy and gains
destination accuracy on changed cells in all five settings (Appendix~\ref{app:accc}).

Pretrained forecasters fare worse -- all three fall below copying, fine-tuned Chronos-Bolt differs
from it on 2 of 281{,}952 cells, and a state-only read-out scores higher overall than
forecast-based decoding. These findings document a negative result and show why the recording process
and evaluation choices must be made explicit.

\section{Task and evaluation}
\label{sec:task}
\paragraph{Cohort and assessment process.}
The export comprises 585{,}791 item entries from 3{,}736 ICU encounters and 39{,}148 questionnaire
days on the surgical ICU of University Medicine Mannheim, Heidelberg University, over roughly six
years; the instrument is the published Ground Truth for Sepsis Questionnaire~\citep{Lindner2022}.
Requiring at least four completed items on both days of a consecutive pair retains \textbf{29{,}715
day-pairs from 2{,}789 encounters}. The export identifies encounters only; person-level identity comes from a
separate table mapping these to 2{,}694 patients. Splitting by encounter -- the convention in the
work we audit -- is \emph{not} patient-disjoint: 24 patients had encounters in both training and
test. Grouping each patient into one fold gives 18{,}688/4{,}708/6{,}319 day-pairs and
1{,}724/431/539 patients, with zero overlap (Appendix~\ref{app:repro}); no result from the encounter
split is carried over.
The 45 targets comprise 25 binary and 20 multiclass items, not all ordinal. Each form covers a fixed
14:00--14:00 interval, is completed by a \emph{single} attending physician per visit, and the
tablet displays that physician's preceding entries, which may contribute to documentation
persistence. A changed value is not deterioration -- ``worsened'' on successive days is
an unchanged label -- so we speak of a \emph{recorded assessment change}.

\begin{wrapfigure}{r}{0.46\linewidth}
\vspace{-\baselineskip}
\centering
\includegraphics[width=\linewidth]{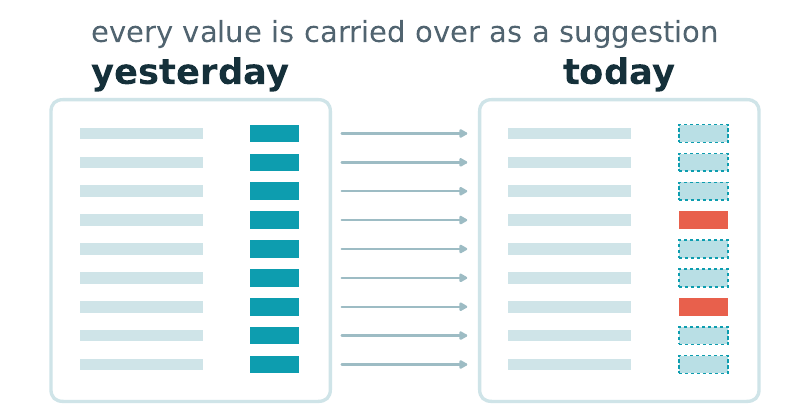}
\caption{How the previous day enters the new form. Each item's previous value is carried into the
form the physician completes, so most rows are confirmed and a few are overridden. This is the
mechanism behind the persistence baseline: it is not only a statistical reference point but part
of the recording process, and the data cannot separate documentation persistence from genuine
clinical stability. Schematic -- the instrument itself is not shown. Two of the nine rows drawn
here differ; in the test data $3.81\%$ of scored cells do.}
\label{fig:carryforward}
\vspace{-\baselineskip}
\end{wrapfigure}

\paragraph{Information available to the predictor.}
For item $j$ the target is $y_{D+1,j}$ and persistence predicts $y_{D,j}$. The form timestamp is always
14:00 -- the interval's reference time, not an entry or signing time -- and
no timestamp recovers when a form was written, so when a physiology-based prediction could have been
issued is not settleable from these data. Consecutive forms are exactly 24.00\,h apart for $98.9\%$ of pairs. Day-pairs are required to fall
on consecutive calendar dates, a separate condition from an exact 24-hour timestamp difference; pairs
failing it are excluded rather than scored.
We therefore report two physiology settings over 24\,h of 107 clinical variables: a \emph{late}
cutoff ending six hours before the target form's reference time, which overlaps the interval that
form assesses, and a \emph{matched} cutoff ending at the \emph{preceding} form's reference time.
Microbiology enters at the time each event became known.

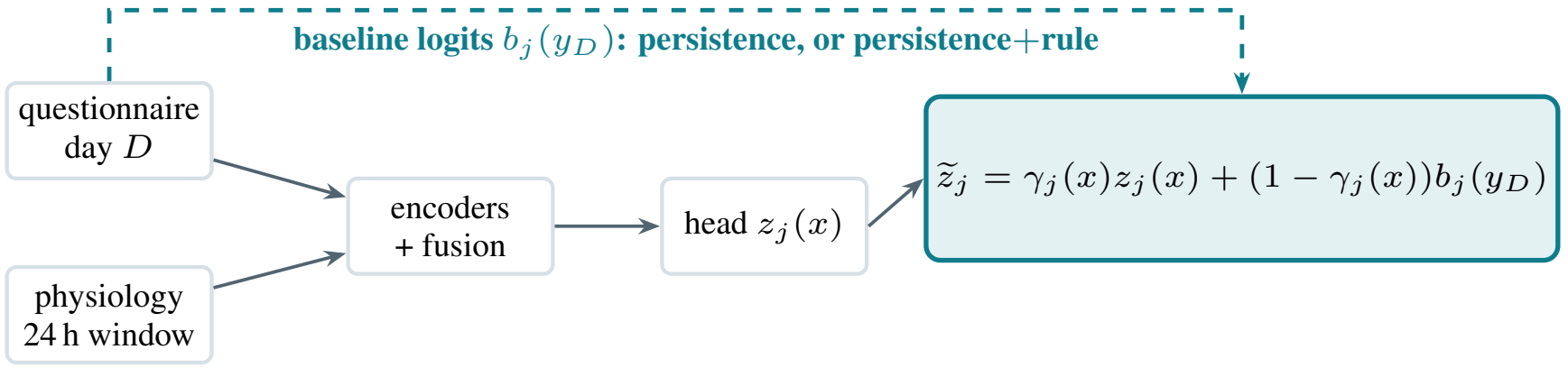
\begin{figure}[t]
\centering
\begin{tikzpicture}[font=\scriptsize,node distance=0pt,
  box/.style={draw=kgline,line width=0.8pt,fill=white,rounded corners=2pt,
              minimum width=15mm,minimum height=7mm,align=center,inner sep=2pt},
  gate/.style={draw=kgteal,line width=1pt,fill=kgsoft,rounded corners=3pt,
               minimum width=30mm,minimum height=12mm,align=center,inner sep=2pt},
  fl/.style={draw=kgmuted,line width=0.7pt,-{Stealth[length=1.6mm]}},
  pers/.style={draw=kgteal,line width=0.9pt,dashed,-{Stealth[length=1.6mm]}}]
\node[box] (q)   at (0,0.6)   {questionnaire\\day $D$};
\node[box] (p)   at (0,-0.75) {physiology\\24\,h window};
\node[box] (enc) at (2.5,-0.1) {encoders\\+ fusion};
\node[box] (head)at (4.8,-0.1) {head $z_j(x)$};
\node[gate] (g)  at (8.3,0.25) {$\widetilde z_j=\gamma_j(x)z_j(x)+(1-\gamma_j(x))b_j(y_D)$};
\draw[fl] (q) -- (enc); \draw[fl] (p) -- (enc);
\draw[fl] (enc) -- (head); \draw[fl] (head.east) -- (g.west);
\draw[pers] (q.north) -- ++(0,0.52) -| (g.north);
% label sits just above the dashed line; both were previously too high and left a visible gap
\node[kgteal,anchor=south,font=\scriptsize\bfseries] at (4.3,1)
  {baseline logits $b_j(y_D)$: persistence, or persistence${+}$rule};
\end{tikzpicture}
\caption{The gate, as evaluated. Encoders and head produce per-item logits $z_j(x)$; a learned
per-item, input-dependent $\gamma_j(x)\in[0,1]$ mixes them \emph{in logit space} (Eq.~\ref{eq:gate}) with a
parameter-free baseline path (dashed) that either copies day $D$ or applies the organ-acuity rule.
Because that path carries no parameters, $\gamma(x)\equiv\mathbf{0}$ reproduces the baseline \emph{exactly}:
the baseline is a selectable special case of the model, not an external competitor, so falling below
it is a statement about what the fitted model chose, not about the hypothesis class.}
\label{fig:gate}
\end{figure}

\paragraph{Baselines and learned predictors.}
The \emph{rule} copies the previous form but maps organ acuity from \newstate{} to \acutestate{}; it
cannot express resolution to \nonestate{}. The \emph{transition matrix} is a per-item first-order
conditional $\arg\max_c P(y_{D+1}{=}c\mid j,y_D)$, fitted on training data without smoothing.
Organ-acuity cells reading \newstate{} on two consecutive days were corrected to \acutestate{} on the
later day; among stable cells the rule can err only on such cells, so after correction its
false-change rate is zero by construction, and every row is retrained and rescored on uncorrected
labels (Appendix~\ref{app:repro}). Supervised models use item-specific cross-entropy. A
gated variant mixes model logits $z_j(x)$ with baseline logits $b_j(y_D)$,
\begin{equation}
\widetilde z_j=\gamma_j(x)\,z_j(x)+\bigl(1-\gamma_j(x)\bigr)b_j(y_D),\qquad \gamma_j(x)\in[0,1],
\label{eq:gate}
\end{equation}
over logits, not labels (Figure~\ref{fig:gate}). The baseline path is persistence, or a
confidence-thresholded per-item transition table which -- verified by enumeration -- installs exactly
the eight organ-acuity \newstate{}$\to$\acutestate{} redirections and so coincides with the rule.

\paragraph{State accuracy and change detection.}
A scored cell is an item observed on \emph{both} days, a change being undefined when day $D$ is
missing; which cells qualify is itself a variable (Section~\ref{sec:results}). Test contains
281{,}952 cells from 539 patients, $r=3.81\%$ changing. With $C=\ind[y_{D+1,j}\ne y_{D,j}]\in\{0,1\}$
we report accuracy, item-macro-F1, $\Accc=P(\hat y=y_{D+1}\mid C=1)$ -- destination accuracy on
changed cells, not change detection -- and $\FCR=P(\hat y\ne y_D\mid C=0)$, related exactly by
\begin{equation}
\Acc-\Acc_{\mathrm{pers}}=r\Accc-(1-r)\FCR,
\label{eq:decomposition}
\end{equation}
so gains on changed cells must exceed errors on stable ones; macro-F1 obeys no such decomposition.
Change scores are reported as AUROC and average precision against an item-identity baseline.
Uncertainty is a paired bootstrap over \emph{patients}; a difference is \emph{established} only when
its interval excludes zero. Full definitions are in Appendix~\ref{app:repro}.

\section{Results}
\label{sec:results}
\begin{table}[t]
\centering
\caption{Test point estimates, all rows from one evaluation pass on the patient-disjoint split, one
mask and one label version. Row groups give the inputs; the \emph{late} cutoff is defined in
Section~\ref{sec:task} and the matched cutoff is in Table~\ref{tab:cutoff}. $\Accc$ is destination
accuracy on changed cells, $\FCR$ the false-change rate on stable cells. Neural rows are five-seed
means; ensembles are in Table~\ref{tab:agg}. Best per column in bold.}

\label{tab:main}
\renewcommand{\arraystretch}{1.06}
\setlength{\tabcolsep}{7pt}
\begin{tabular}{@{}lrrrr@{}}
\toprule
Predictor & Accuracy & Macro-F1 & $\Accc$ & $\FCR$ \\
\midrule
\multicolumn{5}{@{}l}{\emph{Questionnaire input only}}\\
% Produced verbatim by `python exp_fixnn/eval_deploy.py fmts --split_name test
%   --report seedmean --bootstrap 2000`; also written to fmts_table1.tex.
% Do not edit by hand: regenerate. Five-seed means; the ensemble is in Table 3.
% \textbf{}, the \quad indents and the row-group headings below are formatting only: strip them
% and the values equal fmts_table1.tex. Presentation was revised after review (Table 1 was
% reported as hard to parse); no number changed.
\quad Persistence (copy) & 0.9619 & 0.8705 & 0.0000 & 0.0000 \\
\quad Persistence $+$ rule & 0.9632 & 0.8711 & 0.0349 & 0.0000 \\
\quad Transition matrix & \textbf{0.9634} & 0.8681 & 0.0723 & 0.0013 \\
\addlinespace[2pt]
\multicolumn{5}{@{}l}{\emph{Questionnaire $+$ physiology, late cutoff}}\\
\quad GRU without baseline gate & 0.9339 & 0.8033 & \textbf{0.2074} & 0.0373 \\
\quad Gate over persistence & 0.9557 & 0.8687 & 0.0203 & 0.0072 \\
\quad Gate over rule, macro-F1 selected & 0.9609 & \textbf{0.8730} & 0.0443 & 0.0027 \\
\quad Change-weighted objective & 0.9293 & 0.8425 & 0.1253 & 0.0388 \\
\bottomrule
\end{tabular}
\end{table}

\paragraph{A macro-F1 gain, at a cost in accuracy.}
Persistence attains 96.19\% accuracy while predicting no changed cell by construction; the rule and
the transition matrix reach 96.32\% and 96.34\%, recovering 3.49\% and 7.23\% of changed cells
(Table~\ref{tab:main}). The transition matrix's edge over the rule is $+0.000164$
$[-0.000025,+0.000346]$ and is not established. Against that, the macro-F1-selected gate over the
rule reaches 0.8730 macro-F1 and 4.43\% $\Accc$ -- an improvement on item-macro-F1 while losing
accuracy:
$+0.0019$ $[+0.0006,+0.0031]$ macro-F1, against $-0.0022$ $[-0.0026,-0.0019]$ accuracy and a
false-change rate rising from 0.0000 to 0.0027. Gating over plain persistence is worse on every count, including
$\Accc$ $0.0146$ lower \emph{than the rule} $[-0.0204,-0.0086]$: departing from a weaker baseline is
not departing usefully.

\paragraph{The macro-F1 gain depends on task definition.}
Table~\ref{tab:robust} varies the input cutoff, the observation mask and the label version, singly
and together, retraining wherever the training data changes. Ending the physiology window at the previous form's
reference time leaves $+0.0001$ $[-0.0006,+0.0008]$; scoring only explicitly recorded cells leaves
$0.0000$ $[-0.0001,+0.0001]$; under both the gate is significantly \emph{worse} than the
rule, though by a very small margin: $-0.54\times10^{-4}$ $[-0.82,-0.29]\times10^{-4}$. A non-significant interval is not evidence of equality, so the claim
is precise: no macro-F1 advantage is \emph{established} under either alternative. Two effects are
stable across all five settings, each interval excluding zero -- the gate loses accuracy, and it
gains destination accuracy on changed cells, from $+0.0094$ $[+0.0075,+0.0114]$ at the late cutoff
to $+0.0016$ $[+0.0011,+0.0021]$ at the matched one (all five in Appendix~\ref{app:accc}). The trade-off is reproducible; which side
of it the item-macro-F1 summary lands on is not.

\begin{table}[t]
\centering\small
\caption{The gate against the rule under each evaluation choice: differences of five-seed means with
paired patient-clustered intervals, each contrast taken within its own setting. \emph{Late} ends the
physiology window 6\,h before the target form and so inside the interval it describes;
\emph{matched} at the previous form's reference time. \emph{Filled} scores all cells,
\emph{explicit} only those a clinician entered. Entries marked $^{\ast}$ exclude zero; the last row
is scaled by $10^{-4}$. Explicit-mask macro-F1 rests on thin class support (Appendix~\ref{app:mask}).}

\label{tab:robust}
\begin{tabular}{@{}llrr@{}}
\toprule
Cutoff / mask & Labels & Macro-F1 $\Delta$ & Accuracy $\Delta$ \\
\midrule
late / filled       & corrected & $+0.0019$ $[+0.0006,+0.0031]\,^{\ast}$ & $-0.0022$ $[-0.0026,-0.0019]\,^{\ast}$ \\
late / filled       & raw       & $+0.0018$ $[-0.0011,+0.0029]$ & $-0.0033$ $[-0.0036,-0.0028]\,^{\ast}$ \\
matched / filled    & corrected & $+0.0001$ $[-0.0006,+0.0008]$ & $-0.0014$ $[-0.0016,-0.0012]\,^{\ast}$ \\
late / explicit     & corrected & $0.0000$ $[-0.0001,+0.0001]$ & $-0.0005$ $[-0.0007,-0.0003]\,^{\ast}$ \\
matched / explicit  & corrected & $-0.54$ $[-0.82,-0.29]\times10^{-4}\,^{\ast}$ & $-1.29$ $[-1.87,-0.76]\times10^{-4}\,^{\ast}$ \\
\bottomrule
\end{tabular}
\end{table}

\paragraph{Pretrained forecasters stay below copying; fine-tuning converges onto it.}
Chronos-Bolt~\citep{Ansari2024}, Chronos-2~\citep{Ansari2025} and the evaluated MIRA
checkpoint~\citep{Li2025,MIRACheckpoint}, given up to seven days of history, all fall below
persistence (0.9599/0.8542 down to 0.9500/0.8231), agreeing with it on 98.4--99.7\% of cells.
Fine-tuning does not surpass persistence on accuracy or macro-F1: fine-tuned Chronos-Bolt small
differs from the copy on 2 of 281{,}952 scored cells, and MIRA improves markedly on its zero-shot
scores (0.9500/0.8231 to 0.9607/0.8585) while still falling short. In these fine-tuning runs, predictions remained close to persistence.
We claim the narrower reading: these \emph{checkpoints}, under our decoding, do not beat copying
here. A \emph{structural} mismatch with categorical, rarely-changing fields would need a natively
discrete-state control, which we do not evaluate.

\paragraph{A state-only read-out is a strong control.}
These models emit numbers and the targets are categories, so every row above also reflects a decoding
choice: threshold the median, round and clip. Replacing it with a per-item multinomial map from the
nine predicted quantiles to classes -- fitted on train, regularization chosen on validation, forecasts
frozen -- moves $\Accc$ from 1.18\% to 9.51\%, an eightfold change with nothing about the model
altered. The comparison that bounds it is a control. The same read-out given \emph{only the
item's current state}, with no pretrained features, reaches 0.9634/0.8681 with $\Accc$ 7.23\%: higher
on both aggregate metrics, though it recovers fewer changed cells than the quantile read-out
(7.23\% against 9.51\%). It also matches the transition matrix of Table~\ref{tab:main} on every
reported metric, as expected, since a per-item multinomial on a one-hot current state is a
first-order transition table. This is bounded: the forecast-based read-out does not beat a state-only one on the aggregate
metrics, so such an adaptation needs this control before its transition gains are credited to the
model. Adding forecast quantiles to the current-state read-out slightly increases macro-F1 ($+0.0006$
$[+0.0004,+0.0008]$) but reduces changed-cell accuracy ($-0.0036$ $[-0.0051,-0.0022]$); both paired
patient-bootstrap intervals exclude zero (Appendix~\ref{app:fm}).

\paragraph{A pooled change score mostly identifies the question, not the patient.}
A score conditioned on item and current value reaches pooled AUROC 0.8310 on all cells and 0.9166 on
explicit cells, but each item's marginal change frequency alone -- using nothing about the patient --
gives 0.7858 and 0.8640, and within an item it falls to 0.6497 and 0.7469
(Appendix~\ref{app:temporal}): ranking summaries on different populations, not a decomposition into
shares of explained signal.

\section{Discussion and limitations}
\label{sec:discussion}
We find no consistent advantage over strong simple baselines across metrics and settings. The
gate's item-macro-F1 gain over the rule costs accuracy and is not established under either
alternative.

The mask is the sharpest case. The instrument is conditional, so one ``no organ dysfunction'' answer
materializes many zero cells; excluding them raises the change rate from 3.81\% to 5.96\% and lowers
persistence's accuracy from 96.19\% to 94.04\%. The convention is defensible -- the derived values are
logically correct -- but it defines a different evaluation population, and which one is intended
should be stated rather than inherited from a preprocessing default. Appendix~\ref{app:mask} states which we would score.

This single-center study concerns recorded assessments. Displayed prior answers may contribute to
persistence (Figure~\ref{fig:carryforward}), but the data do not separate documentation effects
from clinical stability. 
The instrument changed once during the study period, and one target, \emph{suspicion of
infection}, is not comparable across the two versions (Appendix~\ref{app:repro}).
One severity
item is relative to other ward patients, and label correction removes the stable cells on which the
rule would err. There are 539 test patients and 150 positive new-acuity cells; candidates were
searched on an earlier split with reported patient leakage. We establish neither unpredictability
nor equivalence to the rule. This negative result is specific to the evaluated models and workflow;
the input cutoff, observation mask and decoding rule must be reported and varied.
\label{sec:mainend}

% \clearpage
\bibliographystyle{plainnat}
\bibliography{fmts_references}

% \clearpage
\section*{Acknowledgements}
The authors acknowledge support by the state of Baden-W\"urttemberg through bwHPC and the
German Research Foundation (DFG) through grant INST 35/1597-1 FUGG. 
Marius Fracarolli is funded by the Helmholtz Information \& Data Science School for Health
(HIDSS4Health), Karlsruhe/Heidelberg, Germany. We thank the clinical staff of the intensive care
unit at University Medicine Mannheim for recording the assessments this study analyses, and the
physicians and clinical staff who answered our questions about them.

\appendix
\section{Task specification and reproducibility}
\label{app:repro}
\subsection{Cohort and labels}
Table~\ref{tab:cohort} gives the flow from export to analysis cohort, reproduced by
\texttt{report.py cohort}. The prior version of this work reported 3{,}020 retained encounters; that
figure is not reproducible from any split and is corrected here to 2{,}789 -- the number of
encounters contributing at least one consecutive day-pair. The day-pair count is unchanged.
\begin{table}[h]
\centering
\caption{Cohort flow and split composition. Patient identity comes from the demographics export's
lifetime number, which the questionnaire export does not carry; 15 encounters mapped to more than one
lifetime number and were assigned to the first. Splitting by encounter is not patient-disjoint.}
\label{tab:cohort}
\small
\begin{tabular}{@{}lr@{}}
\toprule
Quantity & Count \\
\midrule
Source item entries / encounters / questionnaire days & 585{,}791 / 3{,}736 / 39{,}148 \\
Questionnaire versions (v1 / v2 / missing) & 213{,}455 / 372{,}334 / 2 \\
Items completed per form, median (IQR) & 13 (9--18) \\
Retained consecutive day-pairs / encounters / patients & 29{,}715 / 2{,}789 / 2{,}694 \\
\midrule
Training day-pairs / encounters / patients & 18{,}688 / 1{,}775 / 1{,}724 \\
Validation day-pairs / encounters / patients & 4{,}708 / 453 / 431 \\
Test day-pairs / encounters / patients & 6{,}319 / 561 / 539 \\
Patient overlap between any two folds & 0 \\
\midrule
Scored test cells, both days observed & 281{,}952 \\
Changed test cells (prevalence) & 10{,}754 (3.81\%) \\
\bottomrule
\end{tabular}
\end{table}

Splitting is by patient, 20\% of patients to test and 20\% of the remainder to validation, seed 42,
not stratified. For reference, the encounter-based split used previously placed 24 patients in both
training and test, 17 in training and validation, and 7 in validation and test.

The 45-item output comprises four judgment items, nine infection-site indicators, and 32
organ-related fields; the last block contains eight presence, eight acuity, eight cause, seven
chronicity, and one capillary-leak field. Infection-site suspicion and confirmation were merged into
binary presence; source-control fields were excluded. Cause categories are nominal: no ordinal
distance is imposed on them anywhere in the evaluation or in the categorical decoding of
Appendix~\ref{app:fm}.
The instrument is published: \citet{Lindner2022} introduce the questionnaire, describe its structure
and report interrater analyses for selected items, and \citet{Schamoni2019} use its labels for sepsis
prediction. What is \emph{not} published, and is specified here instead, is the transformation from
the recorded entries to the 45 modeling targets: the pivot of conditional items into per-organ and
per-site fields, the hierarchy zero-fill, the harmonization of the two questionnaire versions, and
the class encodings. Interrater analysis of selected published items does not extend to those derived
targets.
\paragraph{Questionnaire versions.} The instrument changed once, and the two versions are
\emph{temporally disjoint}: v1 covers 2016-06-01 to 2017-08-27 and v2 2017-08-28 onwards, so v1 is
24.2\% of day-level records and 29.2\% of test day-pairs under the patient-random split. v2 retired
15 item identifiers, none of which are among the 45 targets. One target draws on both codings --
capillary leak, recorded as a sub-field in v1 and a standalone item in v2 -- and is harmonized to a
single binary field. One target is \emph{not} comparable across the boundary: the \emph{suspicion of infection} item admits a \emph{confirmed} class that exists only in v2, where it is the modal value
(v1: 92.3/7.7/0.0\%; v2: 28.9/10.1/61.0\%). Pooling cannot repair this and it inflates any
train-early/test-late contrast on that item, so we report it as a limitation. It does not carry the
headline: recomputing the gate-versus-rule macro-F1 contrast from the saved predictions with that
item dropped from the item average changes it by $+0.0001$.

\paragraph{Ambiguous patient identifiers.} Fifteen of the 9{,}460 mapped encounters carry more than
one lifetime number in the demographics export, and each was assigned to the first. Five of the
fifteen fall in the retained cohort, and for all five the alternative identifier belongs to a patient
in the \emph{same} fold, so no assignment among the recorded alternatives would place an encounter on
both sides of a split. Patient-disjointness therefore does not depend on the tie-break.

\subsection{Prediction time and preprocessing}
Every form's recorded timestamp is 14:00, for $100.0\%$ of 585{,}791 item entries: it is the
reference time of the 14:00--14:00 interval the form describes, not the moment of entry or signing.
No entry or signing time exists in the export, so the interval between a form being written and
becoming available cannot be measured, and no analysis here can establish it. Consecutive forms are
24.00\,h apart for $98.9\%$ of consecutive pairs.

Two physiology cutoffs are therefore reported. The \emph{late} cutoff uses
$[t_{D+1}-30\,\mathrm{h},\,t_{D+1}-6\,\mathrm{h}]$, which is the setting of the earlier work; it ends
six hours before the target form's reference time and so overlaps the interval that form assesses.
The \emph{matched} cutoff uses $[t_{D+1}-48\,\mathrm{h},\,t_{D+1}-24\,\mathrm{h}]$, ending exactly at
the preceding form's reference time. Physiology comprises 107 clinical variables at hourly resolution
with an explicit observed/missing channel per variable; microbiology events are placed at the time
they became known (cultures at collection, results and positivity at result arrival). All
normalization statistics are fitted on training data only.

The label correction rewrites organ-acuity cells reading \newstate{} on two consecutive calendar days
to \acutestate{} on the later day. It is decided against the original series, conditions only on the
preceding day, and is applied to the day-level table before pairing, so a day carries one value
whether it is used as an input or as a target; a day-$D$ input is never corrected using day-$D+1$
information. Corrections affect 27 organ-acuity cells cohort-wide -- 13 training, 6 validation and 8 test under
the patient-disjoint split -- verified by differencing the corrected and uncorrected target arrays on
row-aligned splits. Before correction, 27 of 2{,}470 organ-acuity pairs beginning at \newstate{} were
\newstate{}--\newstate{} (1.09\%); after it, 0 of 2{,}447. The denominator falls because
correcting day $D+1$ removes the \emph{following} pair from those beginning at \newstate{}, and 4 of
the 27 corrected cells had no following pair.

The correction interacts with the rule and this must be read carefully. The rule maps \newstate{} to
\acutestate{}, so the only stable cells on which it can err are those reading \newstate{} on both
days -- exactly the cells the correction rewrites. After correction, 0 of the 2{,}447 organ-acuity
pairs whose first day is \newstate{} remain \newstate{}--\newstate{}, so the rule cannot produce a
false change on these items at all, and its $\FCR$ of zero in Table~\ref{tab:main} is an arithmetic
consequence of the label version, not evidence that \newstate{}--\newstate{} is clinically
impossible. The clinicians' judgment that the form's wording forbids it is the reason for the
correction; it is not independently confirmed by the rule's performance. Because the operation
changes the target, rescoring alone would not measure its effect, so Table~\ref{tab:rawlab} reports
every row retrained \emph{and} rescored on uncorrected labels.

The sensitivity is instructive and runs opposite to the intuition that a correction helps the
corrected-for predictor. The rule's macro-F1 is unchanged (0.8711) and the gated
model's nearly so (0.8730 against 0.8729), though the gate's accuracy does move, 0.9609 to 0.9599. What the correction moves is \emph{persistence}, from
0.8719 on raw labels to 0.8705 on corrected: the \newstate{}--\newstate{} cells were cells persistence
scored correct for free, and removing them costs it exactly those. The consequence is that the
ordering of the two simple baselines depends on the label version -- on raw labels persistence
outscores the rule on macro-F1 ($0.8719$ against $0.8711$), on corrected labels the rule outscores
persistence -- and the paired interval for that difference, $+0.0007$ $[-0.0001,+0.0018]$ on raw
labels, does not exclude zero. On raw labels the rule also makes 8 false changes where on corrected
labels it makes none, confirming that the zero is a property of the label version rather than of the
rule. Any claim that the rule improves on persistence should therefore be read as conditional on an
authorized correction that removes the cells where the rule would be wrong.
Consecutive \newstate{} entries appear to be a recording error: \newstate{} denotes a dysfunction
first entered within the preceding 24 hours, so two in a row cannot both be correct. We proposed the
correction on that basis and the attending physicians who complete the questionnaire accepted it. It
was then applied programmatically to 27 cells, not by case-by-case chart review, so the rule was
adjudicated but individual cells were not. The correction is load-bearing -- it is what makes the
rule's false-change rate zero -- so Table~\ref{tab:rawlab} reports every row retrained and rescored
without it.

\subsection{Models and selection}
All learned rows are GRU encoders over the physiology sequence with concatenation fusion, an
item-wise output head, batch size and learning rate as listed below, dropout, the Adam optimiser, an
epoch cap of 200 and early stopping on validation macro-F1 with patience six. Seeds are 42, 123, 456,
789 and 1337 for every row. The gate mixes logits as in Equation~\ref{eq:gate}; binary baseline
logits are $\pm c$ and multiclass baseline logits assign $c$ to the baseline class and zero
elsewhere, with $c=8$ unless stated. The thresholded transition table used by the \emph{gate over
rule} rows is estimated on the training split only, with minimum support 10 and purity threshold
$\tau=0.5$; a state is redirected only when its modal successor differs from itself. Enumerating its
entries (\texttt{report.py markovmap}) shows it contains precisely the eight organ-acuity
\newstate{}$\to$\acutestate{} redirections, so it is the rule and not a larger table -- a
data-driven baseline given no clinical knowledge recovers the clinicians' rule exactly. This is
verified rather than inferred from the override count that the training log reports. A saturated baseline makes departures expensive, which is an
inductive bias rather than evidence that optimization is solved.

\begin{table}[h]
\centering\small
\caption{Configuration of every learned row. All use a GRU physiology encoder with concatenation
fusion, Adam, an epoch cap of 200, early stopping on validation macro-F1 with patience 6, uniform
loss balance, the both-days mask, and seeds 42/123/456/789/1337. $c$ is the baseline logit scale;
$\lambda$ weights an auxiliary loss that supervises the gate $\gamma_j(x)$ against whether the cell
actually changed, added to the item-wise cross-entropy; $\gamma_j(x)$ is produced by a sigmoid head over
the fused representation, initialized at zero logit. Candidates a--c differ in learning rate, width, depth, dropout, batch size and baseline
scale, as listed; the reported
\emph{gate over rule} row is whichever maximizes validation macro-F1, selected inside the evaluation
script before test is read.}
\label{tab:configs}
\begin{tabular}{@{}llrrrrrr@{}}
\toprule
Row & baseline path & width & layers & ff & lr & batch & drop \\
\midrule
GRU without baseline gate & none & 128 & 2 & 512 & $5\times10^{-4}$ & 64 & 0.05 \\
Gate over persistence & copy, $c=8$, $\lambda=0.5$ & 128 & 2 & 512 & $5\times10^{-4}$ & 64 & 0.05 \\
Gate over rule (a) & table, $c=8$, $\lambda=0.5$ & 64 & 3 & 512 & $2\times10^{-3}$ & 32 & 0.3 \\
Gate over rule (b) & table, $c=8$, $\lambda=0.5$ & 64 & 3 & 512 & $1\times10^{-4}$ & 32 & 0.3 \\
Gate over rule (c) & table, $c=4$, $\lambda=0.5$ & 128 & 2 & 512 & $5\times10^{-4}$ & 64 & 0.05 \\
Change-weighted objective & table, $c=8$, $\lambda=0.5$ & 128 & 2 & 512 & $5\times10^{-4}$ & 64 & 0.05 \\
\bottomrule
\end{tabular}
\end{table}

The change-weighted row adds a reweighting of changed cells by a factor of five in the loss; no other
row reweights. The transition table used as a baseline path is refitted on the training split of
whichever campaign a row belongs to, never carried across campaigns.

\paragraph{Selection protocol, stated exactly.} The hyperparameter \emph{search} reported in the
earlier version of this work was carried out on the encounter-based split and is not repeated here.
What is repeated is the \emph{selection}: the configurations that were validation-optimal in that
search were retrained on the patient-disjoint split with five seeds each, and the configuration
reported in Table~\ref{tab:main} is the one maximizing validation macro-F1 among them. Test was read
once, afterwards, by a single evaluation script. This is a validation-based selection informed by an
earlier exploratory search on a different split; it is not a preregistered protocol and is not
described as one. The full run manifest -- run directory, configuration, seed, selected checkpoint,
input cutoff, label version and aggregation for every row -- is emitted as
\texttt{fmts\_manifest.json} by the evaluation script and released with the code.

\paragraph{Aggregation.} Neural rows in Table~\ref{tab:main} are means over the five seeds, with
standard deviations in the released metrics file. The majority-vote ensemble over the same five seeds
is reported separately in Table~\ref{tab:agg}. Means and ensembles are never combined in one row, and
five seeds are not treated as five independent test cohorts.

\subsection{Metric definitions and uncertainty}
Macro-F1 is the mean over items of the per-item class-macro F1. Each item's class set is fixed from
the training split and held constant on the test split and inside every bootstrap resample, so a
class that happens to be absent from a resample contributes an F1 of zero rather than being dropped
from the average; zero division is scored as zero. Changed-cell macro-F1 repeats that
calculation on cells with $C=1$; it is a class-macro F1 restricted to changed cells, not the F1 of a
binary change detector, and the two are never combined into one claim. Average precision is
\texttt{sklearn.average\_precision\_score}. Micro-F1 equals accuracy for single-label multiclass
targets and is therefore omitted.

Equation~\ref{eq:decomposition} follows by partitioning correct predictions into changed and
unchanged cells: on unchanged cells every departure from persistence is an error. Because both
$\Accc$ and $\FCR$ are reported on the same mask, the identity is checkable from
Table~\ref{tab:main}; Figure~\ref{fig:tradeoff} does not use it, and instead counts departures
directly from the paired per-cell predictions.

Uncertainty is a paired bootstrap over \emph{patients}: 2{,}000 resamples of the 539 test patients
with replacement, all days and items of a sampled patient kept together, and the same resamples used
for every predictor so that differences remain paired. Intervals are 95\% percentile intervals of the
resampled paired difference (2.5th and 97.5th percentiles), uncorrected. Resampling day-pairs, as the earlier version did, preserves the dependence among the 45 items of a
form -- the whole form moves together -- but not the dependence between several day-pairs of the same
patient, which is why the patient is the resampling unit here. Seed variability, patient-sampling uncertainty and ensemble
performance are reported as three separate quantities. We describe a difference as established only
when its interval excludes zero, and use no other significance language.

\subsection{Changed-cell accuracy across settings}
\label{app:accc}
The claim that the gate improves destination accuracy on changed cells in every setting rests on all
five contrasts, not only the two quoted in the main text.
\begin{table}[h]
\centering\small
\caption{$\Accc$ of the selected gate against the rule, under each setting. Differences of five-seed
means with paired patient-clustered bootstrap intervals; every interval excludes zero.}
\label{tab:acccall}
\begin{tabular}{@{}llr@{}}
\toprule
Cutoff / mask & Labels & $\Accc$ difference \\
\midrule
late / filled       & corrected & $+0.0094$ $[+0.0075,+0.0114]$ \\
late / filled       & raw       & $+0.0160$ $[+0.0134,+0.0186]$ \\
matched / filled    & corrected & $+0.0016$ $[+0.0011,+0.0021]$ \\
late / explicit     & corrected & $+0.0109$ $[+0.0085,+0.0134]$ \\
matched / explicit  & corrected & $+0.0004$ $[+0.0001,+0.0006]$ \\
\bottomrule
\end{tabular}
\end{table}

\subsection{The observation mask}
\label{app:mask}
A cell is \emph{explicit} when the clinician entered a value for that field on \emph{both} days of
the pair; the derived cells excluded are those the hierarchy fill materializes from a parent answer.
Transferring that mask onto the main cohort, matched pair by pair on encounter and date, retains
88{,}924 of 281{,}952 scored cells (31.5\%) over the same 6{,}319 day-pairs and 539 patients, and
raises the change rate from 3.81\% to 5.96\%. These rows are a \emph{rescoring}: identical
predictions, targets and models, with only the set of scored cells changed, so nothing is retrained
or reselected.

The explicit subset is a different forecasting population, not the filled one with easy negatives
removed. It retains 5{,}296 of the 10{,}754 changed cells, so it discards 51\% of all transitions
along with the stable cells: requiring an explicit entry on \emph{both} days excludes, in particular,
a transition from a parent-derived negative to a newly documented positive, of which 1{,}207 occur on
the site and organ-presence items. Neither mask is automatically the more clinically valid task, and
we do not argue that the explicit one is; the point is that the comparison changes with the choice.

\paragraph{Which mask we would score.} Asked to choose, we would report the \emph{filled} mask as
the primary population with the explicit mask beside it, and we state that as a judgement rather
than a property of the data. A derived zero is a real clinical assertion -- the clinician denied
the parent condition -- and a forecaster deployed on this instrument would have to produce those
cells too. But the explicit mask is the harder population and the more informative one, and it
discards 51\% of the transitions rather than only easy negatives. Neither alone is sufficient: the
pair is the result, which is why every contrast in this paper is reported within a mask.

\paragraph{Why we do not compare macro-F1 across masks.} Excluding derived cells removes at least one observed class from \textbf{40 of the 45 items} and
leaves \textbf{30} with a single class: the nine infection-site fields fall from two observed classes to one,
because a site is recorded only when ticked, and most organ fields do likewise. Under our
fixed-class-set convention an absent class contributes an F1 of zero, so persistence's item-macro-F1
would read 0.5349; counting only the classes present in each population instead gives 0.9446, because
a single-class item is then trivially perfect. A spread of 0.41 induced by the scoring convention
alone, on the same predictions, is the clearest possible demonstration that the statistic is not
comparable between masks. Neither figure measures predictive
skill. Item-macro-F1 is therefore not comparable between masks and we do not use it that way: the
cross-mask statement is made on accuracy, which is well defined in both populations, and macro-F1 is
retained only for paired comparisons \emph{within} a mask, where both predictors face identical class
support.

\subsection{Aggregation, label sensitivity and temporal transfer}
\begin{table}[h]
\centering\small
\caption{Seed mean $\pm$ sd against the majority-vote ensemble over the same five seeds. The two are
different estimators and are not interchangeable: every number in the main text, every interval and
Figure~\ref{fig:tradeoff} use the seed mean. The ensemble columns appear only here. An earlier version
of this work reported the mean in its tables while bootstrapping the ensemble, which is why the
distinction is made explicitly.}
\label{tab:agg}
\begin{tabular}{@{}lrrrr@{}}
\toprule
Predictor & \multicolumn{2}{c}{Seed mean $\pm$ sd} & \multicolumn{2}{c}{Ensemble} \\
 & Accuracy & Macro-F1 & Accuracy & Macro-F1 \\
\midrule
Gate over persistence             & $0.9557 \pm 0.0006$ & $0.8687 \pm 0.0011$ & 0.9590 & 0.8721 \\
Gate over rule, macro-F1 selected & $0.9609 \pm 0.0008$ & $0.8730 \pm 0.0016$ & 0.9624 & 0.8749 \\
GRU without baseline gate         & $0.9339 \pm 0.0013$ & $0.8033 \pm 0.0012$ & 0.9399 & 0.8154 \\
Change-weighted objective         & $0.9293 \pm 0.0040$ & $0.8425 \pm 0.0063$ & 0.9389 & 0.8540 \\
\bottomrule
\end{tabular}
\end{table}

In these runs the ensemble is better than the seed mean on every row, by more than the seed
standard deviation in each case. We report that as an observation rather than a guarantee of majority
voting; it is not extra evidence about the method, and it is the reason the two are never mixed within a row: reporting an ensemble beside
means would flatter whichever rows happened to be ensembled.

\begin{table}[h]
\centering\small
\caption{Label sensitivity. Every row retrained \emph{and} rescored on uncorrected labels, so the
comparison reflects the training signal as well as the scoring target. On macro-F1 the correction leaves the rule
unchanged and the gated model nearly so, although the gate's accuracy moves by 0.0010; what it moves
substantially is \emph{persistence}, and with it the ordering of the two simple baselines.}
\label{tab:rawlab}
\begin{tabular}{@{}lrrrr@{}}
\toprule
Predictor & \multicolumn{2}{c}{Corrected labels} & \multicolumn{2}{c}{Raw labels} \\
 & Accuracy & Macro-F1 & Accuracy & Macro-F1 \\
\midrule
Persistence                       & 0.9619 & 0.8705 & 0.9619 & 0.8719 \\
Persistence + rule                & 0.9632 & 0.8711 & 0.9632 & 0.8711 \\
Transition matrix                 & 0.9634 & 0.8681 & 0.9633 & 0.8680 \\
Gate over rule, macro-F1 selected & 0.9609 & 0.8730 & 0.9599 & 0.8729 \\
\bottomrule
\end{tabular}
\end{table}

\begin{table}[h]
\centering\small
\caption{Input-cutoff sensitivity. The \emph{late} cutoff ends the physiology window six hours before
the target form's reference time and therefore overlaps the interval that form assesses; the
\emph{matched} cutoff ends it at the preceding form's reference time. Models retrained on each.}
\label{tab:cutoff}
\begin{tabular}{@{}lrrrr@{}}
\toprule
Predictor & \multicolumn{2}{c}{Late cutoff} & \multicolumn{2}{c}{Matched cutoff} \\
 & Accuracy & Macro-F1 & Accuracy & Macro-F1 \\
\midrule
Gate over persistence             & 0.9557 & 0.8687 & 0.9557 & 0.8688 \\
Gate over rule, macro-F1 selected & 0.9609 & 0.8730 & 0.9618 & 0.8712 \\
GRU without baseline gate         & 0.9339 & 0.8033 & 0.9330 & 0.7998 \\
Change-weighted objective         & 0.9293 & 0.8425 & 0.9336 & 0.8460 \\
\bottomrule
\end{tabular}
\end{table}

Against the rule, the macro-F1 difference of the selected gate is $+0.0019$ $[+0.0006,+0.0031]$ at
the late cutoff and $+0.0001$ $[-0.0006,+0.0008]$ at the matched cutoff; destination accuracy on
changed cells goes from $+0.0094$ $[+0.0075,+0.0114]$ to $+0.0016$ $[+0.0011,+0.0021]$. The \emph{macro-F1} advantage is established only at the cutoff that draws on information from
inside the assessed interval, and is not established at the matched cutoff. The changed-cell
improvement is established at both: $+0.0016$ $[+0.0011,+0.0021]$ even at the matched cutoff.

\subsection{Per-item results}
\label{app:peritem}
Table~\ref{tab:peritem} reports all 45 items separately. The aggregate gain is concentrated: of the
$+0.0019$ item-macro-F1 the gate adds over the rule, $+0.0027$ comes from the eight organ
\emph{acute or new} items and $-0.0008$ from the remaining 37. Across all 45 items the gate is
better on 9, worse on 16 and unchanged on 20 at the four decimals shown. Persistence and
persistence $+$ rule coincide on 37 of 45 items, the rule acting only on the new-acuity items.
Forecastability varies widely: \emph{patient's development in last 24\,h} is hardest (macro-F1
$0.5819$ under persistence, with $30.2\%$ of its cells changing), together with the organ
\emph{acute or new} items ($0.628$--$0.646$), while the chronic fields sit at $0.978$--$0.992$ with
under $0.5\%$ of cells changing. The mean over items of each macro-F1 column is the corresponding
aggregate in Table~\ref{tab:main}.

% generated by `python exp_fixnn/analyse_gtsq.py peritem` -- do not hand-edit
% needs \usepackage{longtable} in the preamble
\begingroup\footnotesize
\begin{longtable}{@{}lrrrrrrr@{}}
\caption{Per-item test results, all 45 items. $n$ is the item's scored cells (observed on both days)
and chg\% the share of them that change; macro-F1 is the per-item class-macro F1 whose mean over
items is the aggregate in Table~\ref{tab:main}, so each F1 column averages to that row's published
value. $\Delta$ is gate $-$ rule. $\Accc$ is the gate's destination accuracy on that item's changed
cells. The gate columns are the mean over five seeds.}
\label{tab:peritem}\\
\toprule
Item & $n$ & chg\% & Persist. & Rule & Gate & $\Delta$ & $\Accc$ \\
\midrule
\endfirsthead
\toprule
Item & $n$ & chg\% & Persist. & Rule & Gate & $\Delta$ & $\Accc$ \\
\midrule
\endhead
\bottomrule
\endlastfoot
Daily rating of illness severity & 6,319 & 21.59 & 0.7234 & 0.7234 & 0.7245 & +0.0011 & 0.012 \\
Patient's development in last 24\,h & 6,317 & 30.22 & 0.5819 & 0.5819 & 0.5846 & +0.0026 & 0.016 \\
Working diagnosis & 6,319 & 11.81 & 0.8489 & 0.8489 & 0.8479 & -0.0010 & 0.001 \\
Suspicion of infection & 6,169 & 6.84 & 0.8697 & 0.8697 & 0.8675 & -0.0022 & 0.016 \\
Focus of infection: abdominal & 6,270 & 3.08 & 0.9665 & 0.9665 & 0.9660 & -0.0005 & 0.006 \\
Focus of infection: endocarditis & 6,270 & 0.11 & 0.8701 & 0.8701 & 0.8701 & 0.0000 & 0.000 \\
Focus of infection: blood stream & 6,270 & 1.45 & 0.9515 & 0.9515 & 0.9515 & 0.0000 & 0.000 \\
Focus of infection: skin & 6,270 & 0.62 & 0.9730 & 0.9730 & 0.9730 & 0.0000 & 0.000 \\
Focus of infection: intracranial/meningeal & 6,270 & 0.61 & 0.9700 & 0.9700 & 0.9700 & 0.0000 & 0.000 \\
Focus of infection: catheter & 6,270 & 0.81 & 0.9303 & 0.9303 & 0.9303 & 0.0000 & 0.000 \\
Focus of infection: thoracic & 6,270 & 3.59 & 0.9608 & 0.9608 & 0.9608 & 0.0000 & 0.000 \\
Focus of infection: urogenital & 6,270 & 0.54 & 0.9434 & 0.9434 & 0.9434 & 0.0000 & 0.000 \\
Focus of infection: bone/joint & 6,270 & 0.18 & 0.9852 & 0.9852 & 0.9852 & 0.0000 & 0.000 \\
Capillary leak & 6,245 & 3.96 & 0.9516 & 0.9516 & 0.9507 & -0.0008 & 0.014 \\
Lung: present & 6,263 & 4.06 & 0.9506 & 0.9506 & 0.9504 & -0.0002 & 0.002 \\
Lung: acute or new & 6,263 & 5.75 & 0.6306 & 0.6337 & 0.6794 & +0.0457 & 0.276 \\
Lung: cause & 6,263 & 4.61 & 0.9175 & 0.9175 & 0.9175 & 0.0000 & 0.000 \\
Lung: chronic & 6,263 & 0.32 & 0.9788 & 0.9788 & 0.9788 & 0.0000 & 0.000 \\
Kidney: present & 6,263 & 2.62 & 0.9734 & 0.9734 & 0.9734 & 0.0000 & 0.000 \\
Kidney: acute or new & 6,263 & 3.82 & 0.6456 & 0.6483 & 0.6674 & +0.0191 & 0.254 \\
Kidney: cause & 6,263 & 3.08 & 0.9502 & 0.9502 & 0.9501 & -0.0001 & 0.002 \\
Kidney: chronic & 6,263 & 0.49 & 0.9840 & 0.9840 & 0.9840 & 0.0000 & 0.000 \\
Brain: present & 6,263 & 4.23 & 0.9576 & 0.9576 & 0.9576 & 0.0000 & 0.000 \\
Brain: acute or new & 6,263 & 5.17 & 0.6362 & 0.6396 & 0.6431 & +0.0035 & 0.199 \\
Brain: cause & 6,263 & 4.81 & 0.9344 & 0.9344 & 0.9344 & 0.0000 & 0.000 \\
Brain: chronic & 6,263 & 0.42 & 0.9775 & 0.9775 & 0.9775 & 0.0000 & 0.000 \\
Heart: present & 6,263 & 3.58 & 0.9564 & 0.9564 & 0.9560 & -0.0004 & 0.008 \\
Heart: acute or new & 6,263 & 4.76 & 0.6281 & 0.6332 & 0.6559 & +0.0227 & 0.197 \\
Heart: cause & 6,263 & 3.70 & 0.8510 & 0.8510 & 0.8495 & -0.0015 & 0.012 \\
Heart: chronic & 6,263 & 0.48 & 0.9867 & 0.9867 & 0.9867 & 0.0000 & 0.000 \\
Liver: present & 6,263 & 2.55 & 0.9523 & 0.9523 & 0.9504 & -0.0019 & 0.033 \\
Liver: acute or new & 6,263 & 2.92 & 0.6288 & 0.6329 & 0.6434 & +0.0104 & 0.142 \\
Liver: cause & 6,263 & 2.27 & 0.9314 & 0.9314 & 0.9269 & -0.0045 & 0.021 \\
Liver: chronic & 6,263 & 0.26 & 0.9799 & 0.9799 & 0.9799 & 0.0000 & 0.000 \\
Gastrointestinal: present & 6,263 & 4.14 & 0.9554 & 0.9554 & 0.9554 & 0.0000 & 0.000 \\
Gastrointestinal: acute or new & 6,263 & 4.84 & 0.6356 & 0.6385 & 0.6525 & +0.0140 & 0.160 \\
Gastrointestinal: cause & 6,263 & 4.12 & 0.8595 & 0.8595 & 0.8594 & -0.0001 & 0.001 \\
Gastrointestinal: chronic & 6,263 & 0.10 & 0.9910 & 0.9910 & 0.9910 & 0.0000 & 0.000 \\
Coagulation system: present & 6,263 & 3.21 & 0.9528 & 0.9528 & 0.9524 & -0.0005 & 0.005 \\
Coagulation system: acute or new & 6,263 & 3.91 & 0.6312 & 0.6363 & 0.6462 & +0.0099 & 0.168 \\
Coagulation system: cause & 6,263 & 3.16 & 0.9034 & 0.9034 & 0.9020 & -0.0014 & 0.006 \\
Coagulation system: chronic & 6,263 & 0.14 & 0.9916 & 0.9916 & 0.9916 & 0.0000 & 0.000 \\
Bone marrow: present & 6,263 & 0.93 & 0.9586 & 0.9586 & 0.9562 & -0.0023 & 0.003 \\
Bone marrow: acute or new & 6,263 & 0.96 & 0.6385 & 0.6395 & 0.6362 & -0.0033 & 0.037 \\
Bone marrow: cause & 6,263 & 0.46 & 0.6779 & 0.6779 & 0.6566 & -0.0214 & 0.000 \\
\end{longtable}
\endgroup

\subsection{Ethics and availability}
\paragraph{Institutional review board.} This study analyses retrospectively collected routine-care
data from the intensive care unit of University Medicine Mannheim, Heidelberg University, Germany.
Retrospective analysis was approved by the Medical Ethics Commission II of the Medical Faculty
Mannheim, Heidelberg University, Germany, with waiver of patient consent (reference numbers
2016-800R-MA, 2016-840R-MA and 2023-851).

\begin{figure}[h]
\centering
\includegraphics[width=\linewidth]{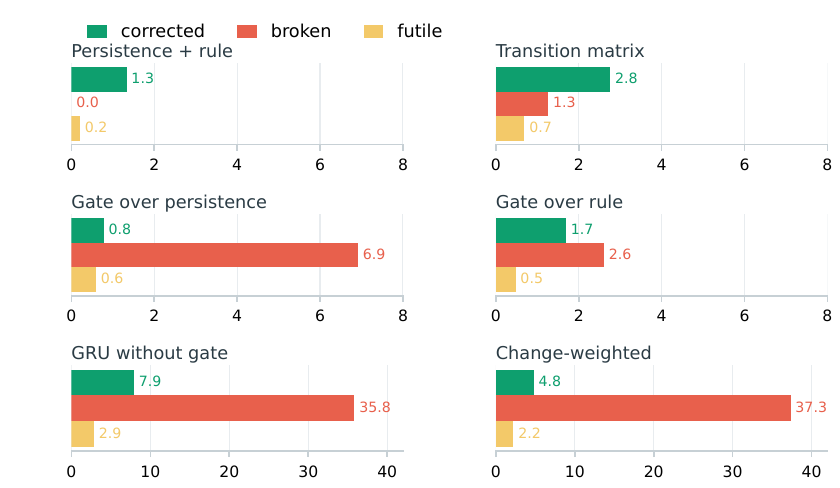}
\caption{Departures from persistence per 1,000 of the 281{,}952 scored test cells, counted
directly from the paired per-cell predictions on the common evaluation mask: corrected
(persistence wrong, model right), broken (persistence right, model wrong), and futile (wrong
either way). These are counts, not reconstructions from aggregate identities. Note the scale:
the two rows without a baseline gate break 35--37 cells per 1{,}000 to correct 5--8.}
\label{fig:tradeoff}
\end{figure}

\paragraph{Data and code availability.} The questionnaire and clinical data are single-center
routine-care records and cannot be made publicly available. The analysis code -- preprocessing,
splitting, training, and the single evaluation script that produces every number in this paper -- is publicly available at
\url{https://github.com/StatNLP/FMTS_2026_When-Copying-Is-Hard-to-Beat} and included as
supplemental material.
The aggregate evaluation artefacts that back Table~\ref{tab:main} and Figure~\ref{fig:tradeoff} --
metrics, patient-clustered intervals, departure counts, and the run manifest recording the input
cutoff, label version, mask, reporting object and seeds of every row -- are written by that script
and included. Per-cell predictions are \emph{not} included: they carry the recorded questionnaire
values and patient identifiers for every scored day, and the persistence row is identical to the
day-$D$ record, so releasing them would release the data itself.

\paragraph{LLM usage.} The main ideas are by the authors.  After completing the initial manuscript, Anthropic's Claude (Opus 4.8, Opus 5 and Opus 5.5), OpenAI's Astra, and paperreview.ai provided feedback and suggestions. Claude also assisted with developing, documenting, and extending the code. No patient-level data from the database were shared with or processed by these tools, in accordance with the applicable data-use agreements.

\section{Additional pretrained-forecaster results}
\label{app:fm}
Histories are up to seven days of questionnaire values (mean 5.51 observed days on test). Chronos-Bolt
processes items individually; Chronos-2 processes them jointly; the MIRA checkpoint is evaluated
itemwise with observation timestamps, padding absent days with NaN rather than a repeated value.
Exact checkpoints, by repository and commit:

\begin{table}[h]
\centering\small
\caption{Evaluated checkpoints. The MIRA weights are the \texttt{MIRA-Mode} open-source
reproduction, whose training corpus differs from the model described in the original
paper~\citep{Li2025}; its pretraining claims are therefore not attributed to these
weights~\citep{MIRACheckpoint}. Chronos-2's support for categorical \emph{covariates} is not
categorical target modeling~\citep{Ansari2025}.}
\label{tab:ckpt}
\begin{tabular}{@{}lll@{}}
\toprule
Row & Repository & Revision \\
\midrule
Chronos-Bolt small & \texttt{amazon/chronos-bolt-small} & \texttt{772f3d25} \\
Chronos-Bolt base & \texttt{amazon/chronos-bolt-base} & \texttt{5d9f166d} \\
Chronos-2 & \texttt{amazon/chronos-2} & \texttt{29ec3766} \\
MIRA & \texttt{MIRA-Mode/MIRA} & \texttt{932caadc} \\
\bottomrule
\end{tabular}
\end{table}

\begin{table}[h]
\centering\small
\caption{Pretrained-forecaster results on the patient-disjoint split and the common mask.
F1$_\Delta$ is changed-cell macro-F1. Agreement is the fraction of scored cells matching
persistence.}
\label{tab:fm}
\begin{tabular}{@{}lrrrrr@{}}
\toprule
Model & Accuracy & Macro-F1 & $\Accc$ & F1$_\Delta$ & Agreement \\
\midrule
Persistence (harness check) & 0.9619 & 0.8705 & 0.0000 & 0.0000 & 100.0\% \\
\midrule
Chronos-Bolt small, zero-shot & 0.9599 & 0.8542 & 0.0070 & 0.0013 & 99.7\% \\
Chronos-Bolt base, zero-shot & 0.9601 & 0.8532 & 0.0118 & 0.0041 & 99.7\% \\
Chronos-2, zero-shot & 0.9570 & 0.8495 & 0.0193 & 0.0110 & 99.3\% \\
MIRA checkpoint, zero-shot & 0.9500 & 0.8231 & 0.0433 & 0.0220 & 98.4\% \\
\midrule
Chronos-Bolt small, fine-tuned & 0.9619 & 0.8705 & 0.0000 & 0.0000 & 100.0\% \\
MIRA checkpoint, fine-tuned & 0.9607 & 0.8585 & 0.0695 & 0.0058 & 99.3\% \\
\midrule
Chronos-Bolt base, read-out on forecast quantiles & 0.9594 & 0.8236 & 0.0951 & 0.0208 & 98.9\% \\
\textit{control:} same read-out, current state only & 0.9634 & 0.8681 & 0.0723 & 0.0204 & 99.5\% \\
\textit{control:} state $+$ forecast quantiles & 0.9633 & 0.8687 & 0.0686 & 0.0203 & 99.6\% \\
\bottomrule
\end{tabular}
\end{table}

\paragraph{Fine-tuning protocol.} Both fine-tuned rows adapt \emph{all} parameters of the released
checkpoint with AdamW (weight decay 0.01), learning rate $10^{-4}$, batch 256, for up to 10 epochs
over 835{,}133 training and 209{,}673 validation item-series, using each implementation's native
objective -- Chronos-Bolt's own quantile loss, MIRA's native loss -- rather than a categorical one.
The epoch is chosen on validation loss; test is never consulted during fine-tuning. All
pretrained-forecaster rows, zero-shot and fine-tuned, are single runs, not seed means; only the
supervised rows of Table~\ref{tab:main} are averaged over five seeds. Fine-tuned Chronos-Bolt small
differs from the copy on exactly 2 of the 281{,}952 scored cells.

\paragraph{Numerical versus categorical decoding.} The rows above decode a numerical forecast by
thresholding the median at 0.5 for binary items and rounding-and-clipping it to the class range for
multiclass items. That imposes a metric on the label space, which is defensible for an ordinal
severity and not for a nominal cause. As a bounded sensitivity check we keep the model's predictive
distribution and map its nine forecast quantiles to class probabilities with a per-item multinomial
logistic regression, fitted on the training split with its regularization strength chosen on
validation. The comparison is decoding against decoding on identical forecasts, so a difference is
attributable to the read-out rather than to a stronger model, and no ordinal distance is imposed on
nominal categories. On identical frozen forecasts, the categorical
read-out moves destination accuracy on changed cells from 1.18\% to \textbf{9.51\%} -- an eightfold
increase -- while macro-F1 falls from 0.8532 to 0.8236 and accuracy is essentially unchanged
(0.9601 to 0.9594). The conclusion that these forecasters barely anticipate transitions is therefore
substantially a statement about collapsing a predictive distribution to its median and rounding it,
not only about the forecasts themselves. Two controls bound what that shows. Giving the same head
only the item's current state, with no pretrained features, reaches 0.9634/0.8681 with $\Accc$
7.23\% -- higher on both aggregate metrics, and numerically the transition matrix of
Table~\ref{tab:main}, which it is. Giving it the state \emph{and} the quantiles reaches 0.9633/0.8687
with $\Accc$ 6.86\%. Against the state-only control these differences are small but resolvable:
macro-F1 $+0.0006$ $[+0.0004,+0.0008]$, $\Accc$ $-0.0036$ $[-0.0051,-0.0022]$, accuracy $0.0000$
$[-0.0001,+0.0000]$, paired and clustered on the 539 test patients. Adding the forecasts to the current state therefore slightly increases
macro-F1 and reduces changed-cell accuracy, both resolvably. We do not claim the forecasts are
uninformative given the state, nor that the net effect is unhelpful: it is a trade-off between the
two metrics, of the same kind the rest of the paper documents, and which side matters depends on
what the forecast is for. The forecast-only read-out, by contrast, does recover more changed cells than the
state-only control ($\Accc$ $+0.0228$ $[+0.0170,+0.0287]$) at a large cost in both aggregate metrics.
All three fitted rows use training labels, are not zero-shot, are not comparable with the rows above
them, and cover Chronos-Bolt base only.

\section{Change-score analysis and temporal holdout}
\label{app:temporal}
The empirical change score is a \emph{fitted} frequency baseline, not a data-free predictor: for each
(item, current value) pair it estimates the training-split frequency of $C=1$. It is fitted on the
training split only, uses a cell only where that cell has at least five training observations, falls
back to the global training change rate for unseen pairs, and applies no smoothing. An item-only
score estimates the corresponding marginal. Within an (item, state) stratum the lookup assigns one
score to every patient, so it cannot rank patients inside a stratum; the within-item figure is the
support-weighted mean of per-item AUROCs over items with both outcomes present in test, and it
measures discrimination between current states, not patient-specific forecasting.

\begin{table}[h]
\centering\small
\caption{Change score on the random and the chronological split. Both are patient-disjoint; the
chronological split places all of a patient's day-pairs in one date range and drops patients who
straddle a cutoff (below). The paired difference has 95\% patient-bootstrap interval
$[+0.0408,+0.0498]$. The rule-covered rows remove
the \newstate{}$\to$\acutestate{} transitions the one-line rule already produces, which are not new
information for a reader who has that rule. A dash marks a decomposition we computed on the random
split only; it is missing, not zero.}
\label{tab:temporal}
\begin{tabular}{@{}lrr@{}}
\toprule
Quantity & Random split & Chronological split \\
\midrule
Scored cells & 281{,}952 & 250{,}682 \\
Change prevalence & 3.81\% & 3.89\% \\
Pooled AUROC (item + state) & 0.8310 & 0.8149 \\
Pooled average precision & 0.2749 & 0.2789 \\
Item identity only, AUROC & 0.7858 & --- \\
Within-item (state only), AUROC & 0.6497 & --- \\
Paired difference, item+state $-$ item only & $+0.0452$ & --- \\
Rule-covered transitions removed, AUROC & 0.8249 & --- \\
Rule-covered transitions removed, AP & 0.2274 & --- \\
\bottomrule
\end{tabular}
\end{table}

\paragraph{Chronological holdout, and what it measures.} Assigning each patient by the date of
their \emph{first} day-pair keeps patients out of two folds but does not separate the folds in time:
a patient first seen in 2016 and readmitted in 2021 carries 2021 data into training. Under that
design our folds spanned 2016-06-01--2022-01-12 (train) against 2020-12-14--2022-03-30 (test), an
overlap of thirteen months. We therefore require every day-pair of a patient to fall on one side of
each cutoff and drop patients who straddle one: 38 of 2{,}694 patients (1.4\%), giving
18{,}738/4{,}432/5{,}577 pairs over \textbf{2016-06-01--2019-10-08}, \textbf{2019-10-11--2020-11-29}
and \textbf{2020-12-02--2022-03-30}, with no overlap and no shared patients. The lookup and all
fitted preprocessing statistics are refitted within the training period.

Strict separation makes the comparison sharper and its interpretation narrower. Training now lies
entirely before the COVID-19 period and test entirely within it, and the cohort changes
substantially: between the training and test folds the prevalence of recorded gastrointestinal
dysfunction rises from 26.5\% to 53.8\% and cerebral dysfunction from 42.3\% to 69.2\%. This row
therefore measures robustness to a large case-mix shift, not to elapsed time alone, and the
questionnaire-version boundary lies inside the training fold, which inflates the apparent shift on
the infection item specifically. What does \emph{not} move is the property the paper is about: the
cell change rate is 3.80\%, 3.99\% and 3.89\% across the three folds, so persistence dominance is not
an artefact of a quieter pre-pandemic cohort.

\begin{table}[h]
\centering\small
\caption{Chronological holdout, point forecasters, five-seed means, on the strictly separated split.
This row is heavily qualified and we do not rest any claim on it. Training lies entirely before the
COVID-19 period and test entirely within it; the cohort is markedly sicker in test (recorded
gastrointestinal dysfunction 26.5\%$\to$53.8\%, cerebral 42.3\%$\to$69.2\%); and the
questionnaire-version boundary falls inside the training period, so one target is coded differently
either side of it. What the comparison shows is robustness to a large case-mix change rather than to
elapsed time. The property the paper concerns does not move: the cell change rate is 3.80\%, 3.99\% and 3.89\%
across the three folds, and against the rule the gated model behaves as on the patient-random split,
$+0.0020$ macro-F1 $[+0.0006,+0.0036]$ at $-0.0030$ accuracy $[-0.0036,-0.0024]$.}
\label{tab:temporalfc}
\begin{tabular}{@{}lrrrr@{}}
\toprule
Predictor & Accuracy & Macro-F1 & $\Accc$ & $\FCR$ \\
\midrule
Persistence & 0.9611 & 0.8641 & 0.0000 & 0.0000 \\
Persistence + rule & 0.9624 & 0.8645 & 0.0324 & 0.0000 \\
Transition matrix & 0.9629 & 0.8617 & 0.0753 & 0.0012 \\
GRU without baseline gate & 0.9222 & 0.7815 & 0.1999 & 0.0486 \\
Gate over persistence & 0.9551 & 0.8627 & 0.0200 & 0.0070 \\
Gate over rule, macro-F1 selected & 0.9594 & 0.8666 & 0.0485 & 0.0037 \\
Change-weighted objective & 0.9348 & 0.8409 & 0.1214 & 0.0323 \\
\bottomrule
\end{tabular}
\end{table}

Neither AUROC nor these retrospective labels establishes the benefit of asking a clinician to review
a field, so alert-budget and decision-curve claims are excluded. The point-forecast and binary-change
metrics are also not combined into a single claim that a model predicts whether but not what: they
are different tasks with different baselines, and we report them separately.

\end{document}